# Metaversal Learning Environments: Measuring, predicting and improving interpersonal effectiveness

**Arjun Nagendran[1*], Scott Compton[2], Bill Follette[3], Artem Golenchenko[1], Anna Compton[1], Jonathan Grizou[4]**

[1] Mursion Inc., San Francisco, CA, USA

[2] Psychiatry and Behavioral Sciences, Duke University, NC, USA

[3] Department of Psychology, College of Science, University of Nevada, Reno, NV, USA

[4] School of Computing Science, University of Glasgow, Glasgow, UK

**\* Correspondence:**
Corresponding Author
arjun.nagendran@gmail.com

## Abstract

Experiential learning has been known to be an engaging and effective modality for personal and professional development. The Metaverse provides ample opportunities for the creation of environments in which such experiential learning can occur. In this work, we introduce a novel architecture that combines Artificial intelligence and Virtual Reality to create a highly immersive and efficient learning experience using avatars. The framework allows us to measure the interpersonal effectiveness of an individual interacting with the avatar. We first present a small pilot study and its results which were used to enhance the framework. We then present a larger study using the enhanced framework to measure, assess, and predict the interpersonal effectiveness of individuals interacting with an avatar. Results reveal that individuals with deficits in their interpersonal effectiveness show a significant improvement in performance after multiple interactions with an avatar. The results also reveal that individuals interact naturally with avatars within this framework, and exhibit similar behavioral traits as they would in the real world. We use this as a basis to analyze the underlying audio and video data streams of individuals during these interactions. Finally, we extract relevant features from these data and present a machine-learning based approach to predict interpersonal effectiveness during human-avatar conversation. We conclude by discussing the implications of these findings to build beneficial applications for the real world.

## 1    Introduction

Over the last decade, advances in computational hardware have accelerated the adoption of virtual reality and artificial intelligence technologies across the workplace. This wave of adoption has led to social media platforms and companies embracing the term "*Metaverse,*" referring to an aspirational seamless boundary between the physical realm and computer-generated realms (i.e., an integrated network of persistent, online virtual or augmented environments (Dionisio, J. D. N., III, W. G. B., & Gilbert, R., 2013)). This adoption has been facilitated by the technological advancements in virtual and augmented reality technology, artificial intelligence, high-speed global connectivity (i.e., internet), and rendering devices with powerful onboard computing. While the term *Metaverse* is commonly associated with virtual or augmented reality experiences, and consequently, the devices capable of realizing them (VR headsets, AR goggles), the intent of *Metaverse,* by definition, is to be all-pervasive and all-consuming. This means that video conferencing, telepresence, virtual worlds and online games, and virtual marketplaces where digital goods can be bought, sold, or traded are all *metaversal* in nature and can be accessed via PCs, laptops, tablets, phones, or other *yet-to-be-built* technologies in the future. With the *Metaverse* hoping to encompass many aspects of our current and future existence, with the aim of enhancing our connectivity and the quality of our shared experiences, the obvious question is how each of us individually interacts with and manifests across these real and virtual worlds. This is where we can draw inspiration from research involving avatars that has already been extensively studied by the research and scientific community.

Avatars: The term "*avatar*" is deep-rooted in hindu mythology and was used to describe the descent of a divine entity from celestial realms to terrestrial regions (Mathew, 2005), often taking a form that was optimized to complete a specific objective. This is synonymous to users in the real world, representing themselves as a digital entity in the virtual world, a concept that seems central to the idea of the *metaverse*. More formally, a *virtual avatar* is described as a perceptible digital representation whose behaviors reflect those executed, typically in real time, by a specific human being (Bailenson & Blascovich, 2004). Individuals who can influence or directly control their virtual counterparts (*avatars*) are referred to as *inhabiters,* although the term *interactor* is also used when the inhabiter is a highly trained professional capable of embodying many different, disparate avatars (Nagendran, Pillat, Hughes, & Welch, 2012). Note, by an extension of these definitions, an *avatar* must always have an objective or a purpose in the virtual world. The purpose could be as simple as providing (or masking) an identity for the inhabiter, or more complex where the characteristics of the *avatar* influence how other avatars and individuals interact with it. If we are to fully embrace the concept of the *metaverse*, it is critical that we understand how humans will behave and interact with real-time digital representations of other humans, either as themselves, or as avatars, in the digital realm. Previous research has shown that avatars can convincingly simulate social scenarios and adaptively steer conversations (Blascovich et al., 2002; Ahn, Fox, & Bailenson, 2012), while eliciting strong levels of social influence (Blascovich, 2002; Fox et al., 2010; Lim & Reeves, 2010). More recently, the use of personalized 3D avatars in a 2D video conferencing context has indicated a higher level of perceived social presence by participants when compared to traditional video (Higgins, Fribourg, & McDonnell, 2021). Researchers have also studied the continuum of avatar appearance and behavioral realism on influencing the valence of human-avatar interactions, and favored higher levels of realism (Bailenson et al,. 2006, Nagendran et al., 2012).

Metaversal Learning Environments (MLEs): The importance of this area of research has implications for various applications that lend themselves to being an integral part of the *metaverse*. For instance, in industries such as customer service, corporate leadership, sales, and healthcare, interpersonal skills training has the primary goal of enabling professionals to deal with challenging situations they may encounter in the real world. The theory behind such training is that exposure to such circumstances will help enable quicker and better decision making. This training can be *metaversal* in nature and has many advantages when compared to traditional role-play based training programs (a comprehensive review can be found in Van Ments, M., 1999; Lane, C., & Rollnick, S., 2007). The use of virtual environments for learning is already being implemented in many professional training programs, particularly to alleviate high costs associated with making mistakes in the real world (Berge, 2008; Herrington, Reeves, & Oliver, 2007). Such *metaversal learning environments (*MLEs*)* are easily scalable and customizable (e.g., outdoors, offices, public spaces, conference rooms, etc.). Providing the ability for individuals to enter and exit virtual environments seamlessly, either as themselves, or in the form of avatars, further amplifies the effectiveness of the training since the avatar manifestation (i.e., age, race, demographic, personality etc.) adds a layer of complexity that can present new learning challenges. Such applications, by virtue of their digital nature, will inherently lend themselves to the collection of high fidelity data streams (e.g., timestamped audio, video, participant tracking, etc.), reduce the need for expensive logistics and planning, and offer the potential to provide quantitative metrics to assess the progress of training.

Despite the volume of research studies centered around human-avatar representation and perception, there are currently no available frameworks, to our knowledge, that can measure the effectiveness and the outcome of interpersonal interactions between avatars and humans. In the absence of a well defined framework, it is difficult to assess the value of such *MLEs*. By defining a framework where the nature of human-avatar interaction can be measured, we can track the progress and improvement of individuals who make use of such *MLEs*. In addition, such data will allow us to systematically evolve avatar representations in the future by injecting nuanced and high fidelity behaviors that capture both the intent and emotional states of inhabiters, thereby creating more engaging modalities of learning.

Interpersonal Effectiveness: One such *MLE* is the use of avatars to assess and shape interpersonally effective behaviors around a specific context or task. Exactly what constitutes the construct of interpersonal effectiveness is not universally agreed upon (Phillips et al., 2016). For our purposes two important aspects of the definition include: (1) who determines whether a behavior is effective ? and (2) does the behavior have to meet a topographically specified criteria to be deemed effective ? We can address the first aspect with the following example: If Person A must convince Person B about their views surrounding a specific topic, it is the final opinion of Person B that matters with regards to whether or not Person A was successful in doing so. It does not matter whether Person A believes they did a good job if it does not correspond to Person B's assessment of the same. With respect to the second aspect, it is not necessary to engage in specific behaviors in order to accomplish a particular goal, nor will following a specified set of behaviors guarantee a successful outcome. While exhibiting theoretically learned behaviors may be useful, doing so does not guarantee success. And finally, for Person A to be interpersonally effective, Person B must be convinced of Person A's view without coercion or merely deferring to authority. For example, when a commanding officer orders an enlisted person to do a task, compliance does not indicate interpersonal effectiveness because no other outcome is free from aversive consequences.

Learning Interpersonal Effectiveness: Based on the above, we believe that interpersonally effective behavior is learned. Such behaviors are selected based on the consequence they have for the person emitting them. Behaviors that increase the likelihood of achieving a goal are strengthened. Learning is most efficient when a behavior is followed by immediate consequences rather than delayed outcomes. Though one might predict improvement over multiple interactions with an avatar, one would predict faster learning if consequences in that moment were noticeable by the individual. Additionally, one might predict that the largest improvements in interpersonal effectiveness would be seen in individuals who have sufficient room for growth in their behavioral repertoire. There are processes that are likely to be associated with interpersonal effectiveness. The first is that during the course of the interaction, the person assessing the interaction generally finds the conversation to be positive. The second is when the person seeking a goal recognizes that the interaction is not being received positively, they alter their behavior to return to a positive valence. How that is done is up to the person trying to achieve a goal. If one cannot recognize that the person they are interacting with is not having a positive experience and cannot emit an alternative behavior if needed, the likelihood of achieving a favorable outcome is diminished.

In this manuscript, we leverage the power of *MLEs* to study the phenomenon of interpersonal effectiveness during human-avatar interaction. To create our measurement framework, we draw inspiration from literature in the social sciences on social effectiveness that indicates that success during any interpersonal interaction is a function of the continuous conversational impact that each person is having on the other. In other words, how people feel is influenced by what is being said (i.e., verbal) or done (i.e., nonverbal) in the moment, with overall positive mutual feelings leading to better conversational outcomes. We extend this paradigm and develop a framework that can help answer the following research questions:

1. Can an individual's interpersonal effectiveness be measured using avatars ?
2. Do an individual's actions (verbal and non-verbal) throughout a conversation influence the eventual outcome of that conversation, measured from the avatar's perspective ?
3. Does an individual's interpersonal effectiveness improve over repeated interactions with avatars in MLEs ?
4. Can the data captured in MLEs be useful in measuring or predicting the interpersonal effectiveness of an individual interacting with an avatar ?

To answer these research questions, we introduce a novel framework to measure interpersonal effectiveness in an MLE. We first conducted a pilot study to better understand how individuals in the real world naturally interact with avatars in the virtual world. The results from this pilot study informed the development of a novel set of algorithms that allowed a user to inhabit their avatar with very little cognitive load. This served as the foundation for our main study involving human participants interacting with an avatar across multiple interactions. The emphasis was on determining whether an individual's interpersonal effectiveness (as measured from the inhabiter's perspective) was correlated to their likelihood of success surrounding that topic of conversation. In addition, we wanted to measure and analyze any change in the interpersonal effectiveness of people repeatedly interacting with the avatars. Finally, we applied machine learning algorithms to analyze the data collected during these interactions in the MLE, to assess if the underlying datastreams, such as facial expressions or vocal patterns, have any predictive power with respect to the outcome of the interaction. A summary of the main contributions of our manuscript is below:

**Contributions:**

1. We present the components of a low-cognitive load system to control avatars in MLEs using audio-based Artificial Intelligence (AI).
2. We present a novel measurement framework to evaluate an individual's performance during a conversation with an avatar that has two components: (a) a moment-by-moment rating of the effectiveness of an individual and (b) an overall measure of their performance.
3. We present results from a research study that uses data from this novel framework to analyze participants' interpersonal effectiveness across multiple interactions with an avatar.
4. We present a framework that uses machine learning to analyze this data and predict the individual's likelihood of success during conversations with the avatar.

In the following section, we describe related work in the area of human-avatar interaction and interpersonal effectiveness that are both instrumental to the development of our novel framework.

## 2 Related Work

### 2.1 Avatar-mediated interaction

The *metaverse* is likely to contribute to an increase in the number of human-avatar interactions for the purpose of training complex behaviors in the real world. Users have reported various advantages to interacting with avatars when compared to live role playing, including anonymity during real-time social events, synchronous communication, ease of creating new contexts, and increased connectivity between individuals from diverse geographic regions (Green-Hamann et al., 2011). Further, avatar-mediated communication has the potential to compensate for missing nonverbal cues and provides rich visual information to its users (Walther, 2007). However, despite these benefits, there has been debate on whether human-avatar social interactions result in similar social responses as seen in human interaction (Roth, 2019) and whether the behaviors learned during human-avatar interactions generalize to the real-world. For example, avatars may be perceived in a different manner when compared to human interactional partners (de Borst & de Gelder, 2015), perhaps due to the feelings of eeriness they may evoke in users (Schwind & Jager, 2016).

Research has found that a key component for successful avatar-mediated interaction is whether an avatar can be perceived as authentic, or real. Work done by Johnsen and colleagues (2005) showed that to be successful, avatars must be perceived as authentic when leading training sessions for physicians. It was suggested that the reduced expressiveness of avatars contributed to a decreased perception of avatar authenticity (Raij et al., 2007). However, work done by Roth and colleagues (2019) demonstrated that avatars can be perceived as authentic, sometimes more authentic than pre-recorded videos of humans, suggesting that avatar-mediated interactions can be successfully applied to different social contexts. It was hypothesized that authenticity served as a catalyst for affective empathy, or the ability to recognize the emotional state of another person, and cognitive empathy, or the ability to take the perspective of someone else. Similarly, research has shown that avatars are able to elicit similar emotional responses in a conversational partner when compared to interacting with a live human (O'Rourke, S. R., et.al. 2020). More realistic avatars have been shown to be perceived as more human-like and have the potential to evoke feelings of virtual body ownership (e.g., when one accepts the virtual body as their own; Latoschik et al., 2017).

Another key component to effective communication is establishing rapport, which can arise when participants exhibit mutual attention, positive behaviors (i.e., smiling, head nodding), and coordination (i.e., leaning forward, body positioning) (Tickle-Degnan & Rosenthal, 1990). Research has found that avatars are capable of establishing rapport with their conversational partners by demonstrating these behaviors during interactions (Wang & Gratch, 2010, Roth et.al., 2019). In addition, avatars are able to represent human facial expressions, including happiness, disgust, surprise, anger, fear, and puzzlement, allowing them to be perceived as authentic and empathetic (Rizzo et al., 2001). When successful, avatar-mediated communication can be applied to a variety of situations. For example, a framework called AMITIES (Nagendran, A., et al., 2014) has been shown to be used effectively for teaching and training necessary job skills (Hughes, C. E., et al., 2015). Elements of this framework have been enhanced to create the software utilized in this research study. Another avatar-mediated interactive system called "Exceptionally Social" has shown success in promoting social skills in children with autism by providing them with an opportunity to practice conversations in a safe and diverse virtual environment (Nojavanasghari et al., 2017). Similarly, another automated social skills trainer has been used to enhance social skills training in individuals with autism by adding audiovisual feedback to the interaction (Tanaka et al., 2017). The benefits of avatar-mediated skills training has also been seen in individuals diagnosed with social communication disorders, where affective avatars were able to elicit empathetic reactions in participants and enhance skills training (Johnson et al., 2016).

2.2  Interpersonal Effectiveness and Emotional Intelligence

Humans are social and predisposed to form and maintain close interpersonal relationships. In fact, failure to form attachments has been linked to negative health, poor social adjustment, and overall well-being outcomes (Baumeister & Leary, 2007). It remains to be established whether or not such relationship forming is likely to occur with avatars, and more importantly if behaviors learned in such *MLEs* translate to real world interpersonal interactions. That question aside, an important factor that goes into forming and maintaining interpersonal relationships is the ability to recognize the impact one is having on another person. This ability, which we refer to as interpersonal effectiveness, is closely associated with conventional definitions of Emotional Intelligence (Kunnanatt, J. T., 2004) or Self-Awareness. The term emotional intelligence (EI) was first defined by Salovey and Mayer (1990) and described as the "*ability to perceive accurately, appraise and express emotions; the ability to understand and regulate emotions and to use this information to guide one's thinking and actions.*" Research has examined the relationship between EI and authentic leadership, which is composed of self-awareness, relational transparency, internalized moral perspective, and balanced processing (Walumbaw et al., 2008). Work by Miao, Humphrey and Qian (2018) has suggested that there is a positive association between EI and authentic leadership, which has practical implications for the workplace as it allows leaders to achieve desirable outcomes across organizational levels. Further, work by Gardner and colleagues (2009) found that leaders who score high on the emotion perception branch of EI are better at perceiving others' emotions, allowing them to create empathetic bounds and be a more authentic leader. Similarly, leaders who are aware of their own emotions and understand the impact they have on others (e.g., self-awareness) are perceived by their subordinates as being more effective leaders (Butler, Kwantes & Boglarsky, 2012). Further support for the importance of EI in effective leadership can be seen in the work done by Braddy and colleagues (2019) who found that women who overrated their leadership behaviors were perceived as being more risky and less effective as a leader by their supervisors when compared to women who underrated their leadership behaviors. These findings suggest that it is particularly important to be able to recognize one's interpersonal effectiveness when working in both personal and professional settings and leadership roles.

In summary, previous research has demonstrated the benefits of avatar-mediated interactions. Researchers have used this to their advantage, creating high fidelity, safe, immersive learning environments that produce similar results to interpersonal interactions in the real world, while mitigating the cost associated with committing mistakes that could negatively impact interpersonal relationships. However, there are not many frameworks that help us understand human behavior or measure conversational success and improvement in performance when repeatedly interacting with avatars (or other humans). We attempt to address this issue by leveraging the power of *MLEs* and creating a framework where avatars can help measure, predict and improve an individual's interpersonal effectiveness, described in detail in the following sections.

## 3  Pilot Study

We first conducted a pilot study to develop and refine the software interface that allowed humans to interact with avatars naturally in an *MLE*. The software leverages Virtual Reality and Artificial Intelligence routines described in detail in Sections 3.2 and 4.2 of this manuscript. The pilot was intended to test the data collection and processing algorithms, enhance the user interface, and develop an interpersonal effectiveness measurement framework during human - avatar interaction that could be validated in a follow-up study.

### 3.1  Participants

The study was conducted at Mursion, a San Francisco based professional training company that delivers high fidelity avatar-based simulations across multiple industries. An e-mail was sent to all employees at Mursion describing the study and asking for volunteers to participate. Care was taken to ensure that employees understood that participation was voluntary and that not participating would in any way impact their employment status at Mursion. Fifty-nine (n=59) out of 75 employees agreed to participate in the pilot project. There were 32 female participants and 27 male participants from various departments of the company, and at various career levels within the organization, providing a reasonably representative sample of the population of employees at Mursion. Participants were provided detailed information about the purpose of the study beforehand and provided verbal informed consent prior to participating. Consent was recorded in runtime through the VR software interface. All participants were allowed to participate in the study independent of whether or not they were willing to provide the video recording consent, which contained data needed for analysis.

### 3.2  VR Software Architecture

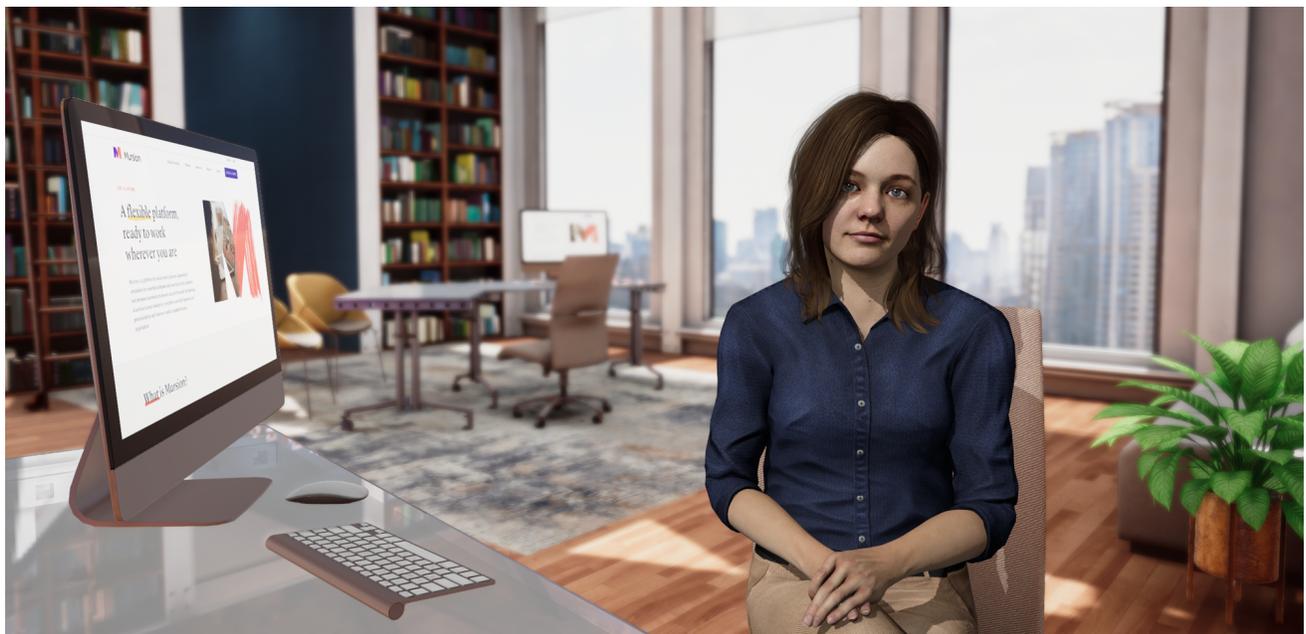

*Figure 1: A metaversal learning environment (MLE) can be used to simulate a challenging conversation with a colleague, represented by an avatar in a workplace setting.*

The VR software used for the study was built using the game development engine Unity (Haas, J. K., 2014). Photogrammetry was used to generate the avatars. The virtual environments used were modeled in Maya (Autodesk, INC., 2019) and imported into the rendering engine. The software had two synchronized peer to peer networked components; an *authoritative* component that was used by an inhabiter to control the avatar and a *non-authoritative* component that was used by participants to interact with the avatar. An illustrative render of the MLE is shown in Figure 1 below and the software architecture diagram along with the collected data streams is shown in Section 4.5.2.

3.2.1 *Authoritative Component:* The authoritative instance was controlled by an *inhabiter* and responsible for creating a secure networked room using an adapted native version of the Real Time Communication (RTC) for Web protocol (Rescorla E, 2013). An Xbox controller allowed the *inhabiter* to select one of several facial gestures and body language in real-time to express themselves via their avatar (Zelenin, A., Kelly, B. D., & Nagendran, A., 2019). A microphone with headset was used for communication. A continuous 10ms buffer of audio data was processed in real-time to predict the probability of the segment containing one of the 5 different phonemes, /a/, /e/, /i/, /o/, or /u/. In addition to this, the RMS value in the audio segment was used to determine the power in the vocal burst and combined with the predicted phoneme probability to perform lip synchronization of the avatar being controlled by the *inhabiter*.

The *inhabiter* was also given a keyboard interface and trained via a certification process to use the keyboard to assess the moment-by-moment interpersonal performance of the participant during the interaction, with respect to the pre-specified outcome of the interaction. The interface allowed for the selection of three impact states: positive, neutral, or negative. If the *inhabiter* thought that the performance of the participant was positive at any given moment, he or she would indicate this rating using the keyboard. This rating would continue as positive until the inhabiter felt that the performance of the participant was either neutral or negative, at which point they would indicate a change in their moment-by-moment rating (see IMPACT, Continuous Rating of Interpersonal Performance, Section 3.5.2). At the end of the conversation, the *inhabiter* also rated whether the participant achieved the outcome of the interaction using the SURVEY (see Post-conversation Rating Of Success, Section 3.5.1) launched via the software interface. Throughout the interaction, the *inhabiter* was able to receive a video and audio feed of participants who entered the MLE, provided the participant granted them the necessary device permissions.

3.2.2 *Non-authoritative Component:* The non-authoritative instance operated on a wide range of devices by participants who wished to interact with the avatars in the virtual world. The system was designed to be compatible with virtual reality headsets, augmented reality eyewear, smartphones, tablets, and standard desktops or laptops. For the purposes of this study, participants used a laptop to enter the secure RTC room created by the authoritative instance using their emails and a pre-chosen passphrase that went through authentication. Once logged in to the room, participants could provide permissions for the transmission of their audio and video feed so that bi-directional communication between them and the avatar was enabled. A synchronized rendering of the avatar controlled by the *inhabiter* through the authoritative instance was realized on the non-authoritative instance using reliable and unreliable ordered and unordered packets, sent and received using a combination of TCP and UDP protocols. This interface allowed the participants to interact with the avatar in a peer-peer manner in the same way as a video conferencing system. While they were unable to see the actual *inhabiter*, they were able to see and hear the inhabiter's intent manifested through the avatar on screen. This architecture facilitated participants in the real world to have interactions with avatars in the MLE, in turn providing us with the necessary data to study interpersonal effectiveness in this specific context.

## 3.3 Experimental Design & Protocol

For the pilot study, a single topic of conversation was chosen for all participants. Participants were emailed a link to an introductory video that explained the concept of interpersonal effectiveness, and asked to utilize this framework during their interaction with the avatar. A link to the narrative of the video can be found [here](). The chosen topic of conversation (also referred to interchangeably as *Scenario* in the rest of the manuscript) involved addressing a communication style difference between the participants and the avatar. The description of this *Scenario* was available for all participants to read prior to their interaction in the MLE and can be found [here]().

In the software, participants first interacted with a host avatar so they could familiarize themselves with the interface and also ask any questions they may have about the social interaction that was about to occur. The host avatar also helped troubleshoot any issues such as sound, rendering, or network latency so the data from the actual interaction was valid and usable for analysis. Following the initial interaction with the host, participants proceeded to interact with a different avatar that represented the *inhabiter* in the specific *Scenario*. Artificial Intelligence technologies, including voice-morphing were used so that a single *inhabiter* could play the role of both the host avatar as well as the avatar used for the social interaction scenario in the software. The description of these technologies is described in detail in previous literature (Nagendran et al., 2012; Zelenin, A., Kelly, B. D., & Nagendran, A., 2019). Four *inhabiters* controlled the authoritative instance and avatars that all participants interacted with during the pilot study.

## 3.4 Data Collection and Measurement Framework

During the interaction between the participants and the avatar, the authoritative instance described in Section 3.2.1 collected various data streams and housed them in a local repository, including the primary outcomes described in detail below. Both instances of the software (see Section 3.2) were also designed to separate the audio and video streams of the *inhabiter* and the participant at the source, and combine them in a synchronized manner into a single container. In addition, the rendered avatar video as well as the gestures and facial expressions of the avatar were recorded in synchrony with the audio and video streams. An interpersonal effectiveness measurement framework was then created to analyze all the collected data during the *Scenario*.

3.4.1 <u>Primary Outcomes</u>

*3.4.1.1 <u>Post-conversation Survey Rating of Success (SURVEY)</u>*

Prior to each conversation, participants read a brief description of the *Scenario* for backstory and context. The description also presented the participant with a pre-specified outcome (i.e., goal) to be achieved by the end of the conversation. Participants were not given instructions or suggestions on how to achieve the pre-specified outcome. They were allowed to use any strategy of their choosing. Following each conversation, the inhabiter and the participant were asked to indicate how likely this outcome was achieved using the SURVEY-I or SURVEY-P, respectively. For the inhabiter, the SURVEY-I consisted of the average of the following two items, each rated on a 1-to-10 Likert-scale:

1. Following this conversation, indicate how likely you are to try <insert outcome>, where 1 = "Extremely Unlikely" and 10 = "Extremely Likely."
2. In coming to a decision, did you feel your views were considered or dismissed, where 1 = "Extremely Dismissed" and 10 = "Extremely Considered."

For participants, the SURVEY-P consisted of a single item rated on 1-to-10 Likert-scale:

1. Based on the conversation you just had, indicate how likely <Name of avatar> will try <insert outcome>, where 1 = "Extremely Unlikely" and 10 = "Extremely Likely."

The SURVEY-I consisted of the average of two items, rather than a single item, to correct for the possibility that a participant may use coercion as a strategy to achieve the goal (e.g., "you must do <insert goal> because I'm your superior and I'm telling you to do so!"). This data was also recorded as a binary indicator of conversation outcome (success/failure) with SURVEY scores $\geq 7$ indicating a "successful" conversation and scores $< 7$ indicating an "unsuccessful" conversation.

### 3.4.1.2 *Continuous Rating of Interpersonal Effectiveness (IMPACT)*

This measure captures the continuous impact the learner was having on the inhabiter (and consequently, the avatar) by taking into account the behavior of the learner, both verbal and non-verbal, during the conversation with respect to the pre-specified outcome of that *Scenario*. During each conversation, the inhabiter provided a continuous rating of the performance of each participant using a three-level ordinal scale (positive, neutral, negative) captured via the mechanics described in Section 3.2.1. This process resulted in a continuous stream of data that reflected the participant's performance throughout the conversation from the perspective of the inhabiter. This data was then processed to generate an overall IMPACT score, with higher values reflecting better interpersonal effectiveness.

### 3.4.2 *Audio streams*

Both the participant and the inhabiter were required to use a headphone with a microphone during the interaction to record clean audio data. Separate audio tracks were created for each on a peer connection event and marked with unique IDs. The data was recorded as a mono channel ranging between 16KHz and 48KHz, depending on the audio device used by the inhabiter or participant. The open source Google voice activity detection algorithm (GoogleWebRTC, 2016) was used to identify voiced and unvoiced segments of the audio. Section 4.5.3.2 describes additional features that were extracted from this data which was time-stamped to be synchronized with the IMPACT (see Section 3.2.1) datastream.

### 3.4.3 *Video streams*

The video feed from both the participant as well as the inhabiter was captured using webcams on each instance of the software. As with the audio, the video was created on separate channels with unique IDs and a three-channel master container was recorded. The recording therefore consisted of the two individual video channels (for participant and inhabiter) and a master channel which contained a stitched video of the virtual environment and avatar, alongside a video of the participant, which was then used for playback and review. The video was captured at a fixed frame rate ranging between 15 and 30 frames per second depending on the local device configuration. Synchronization was performed to the rendering frame rate which was dependent on the local system performance to ensure integrity of the stitched master channel for video review and playback. An open source deep learning algorithm (Baltrusaitis, T., Zadeh, A., Lim, Y. C., & Morency, L. P., 2018) was used to detect and process the facial data as well as the head orientation of the participant and the inhabiter during the entire social interaction. Additional features extracted are detailed in Section 4.5.3.1. This data was time-stamped and stored so it could be synchronized with the IMPACT (see Section 3.2.1) datastream.

### 3.4.4 Post Interaction Questionnaire

Following the interaction, we presented several questions to the participants. The goal of these questions was to help inform a better design where we could truly measure the interpersonal performance of individuals interacting with avatars in *metaversal learning environments*. The questions were mostly open-ended in nature but a few contained multiple choices. The full questionnaire can be found [here](#).

## 3.5 Results

The data collected during the pilot study was analyzed to help inform a more comprehensive follow-on study. Since all data from the interactions was recorded, we benefited from an extended debrief with the inhabiters, analyzing the recorded data, and watching all the videos of the interactions. Three independent individuals were asked to watch the videos and debrief with the inhabiters while recording their observations. In addition, we correlated timestamped logs of the avatar actions with the inhabiter's continuous impact rating data. Our findings are summarized below:

3.5.1 *Measurement and Interaction Framework:* From the recorded observations and the correlation analysis we learned that the body language and facial expressions of the avatar, controlled by the inhabiter via the Xbox, were not aligned with the IMPACT data and events of interest data that were recorded by the inhabiter via the keyboard interface. In addition, the cognitive load on the inhabiter to use the Xbox and the keyboard interfaces simultaneously was high. We overcame these limitations by using a novel audio-based AI algorithm that reduced the cognitive load of inhabiters (described in Section 4.2).

3.5.2 *Post-Interaction Questionnaire:* Results of the survey revealed that more than 95% of the participants were willing to watch a video recording of their own performance, in order to help them improve their interpersonal effectiveness. In addition, participants showed a clear preference for an absolute measurement scale to indicate their interpersonal effectiveness during the interaction.

3.5.3 *Open-Ended Findings:* The post survey questionnaires also included open ended responses to help us develop a better interaction experience in the MLE. There were three major themes in the responses, illustrated by some select comments:

(i) Participants wanted to view moments in the interaction where they did something good or something bad.

> *"The feature of tagging moments where I initiated and responded in savvy ways with Jenna and moments where I missed her cues or went in the wrong direction would be really helpful."*

> *"I'd suggest providing participants with a number of "suggested markers" i.e. "Good moment - empathy", "good moment - listening", "bad moment - prescribing" etc. It would help with consistency and data metrics."*

(ii) Participants wanted to understand whether or not they improved in achieving the outcome of the conversation on a scale that was not just binary.

> *"No/Yes needs options in between--particularly for those who were "getting there" to feel some success and better acknowledge what did and did not go well."*

> *"I like the idea of someone's score being tracked against their own performances so they can clearly see improvement that has been made."*

(iii) Participants were keen on knowing how they compared to other participants who also had the same conversation.

> *"I think the score without detail (even if it's just how it measures up against your peers or folks doing similar scenarios) would be great"*

> *"I could see the comparison scores being helpful within an organization, IF it is positioned as a way to gain overall relational improvement aligned with the company's culture, mission and values."*

Guided by the above findings, we proceeded to refine the software interface and algorithms, so we could conduct a comprehensive study to test our hypotheses. This is outlined in detail in the section that follows.

## 4    Main Study

### 4.1    Primary Objectives

The primary objectives of the main study were: (1) to evaluate the association between continuous ratings of interpersonal effectiveness (IMPACT) collected during conversations and post conversation ratings of success (SURVEY); (2) to evaluate improvement in ratings of interpersonal effectiveness (IMPACT, SURVEY) between *Scenario 1* (screening conversation) and *Scenario 4* (post-assessment conversation) after four interpersonally challenging conversations with avatars ; (3) to explore whether improvement in ratings of interpersonal effectiveness (IMPACT, SURVEY) differ between participants rated as "successful" or "unsuccessful" following the initial screening conversation (*Scenario 1*); and (4) to evaluate whether video and audio data streams collected during conversations in the MLE can be used to predict post-conversation success probability (SURVEY scores).

4.1.1 <u>Hypotheses</u>

Based on our primary objectives, our research hypotheses are presented below:

1. Continuous ratings of interpersonal effectiveness (IMPACT) and post-conversation ratings of success (SURVEY) will be positively correlated.
2. Participants' interpersonal effectiveness scores (IMPACT) will significantly improve between *Scenario 1* and *Scenario 4* and
3. Those participants rated as "unsuccessful" at the end of *Scenario* 1 (failed to achieve the outcome of conversation 1) will show more improvement by *Scenario 4* than those rated as "successful." *by Scenario 4*
4. Verbal and nonverbal behaviors of participants during a *Scenario* will predict interpersonal effectiveness (SURVEY).

To test our research hypotheses, we designed a comprehensive study (see Section 4.4) where participants could interact with the same avatar across four different scenarios centered around challenging conversations in professional settings. We used the knowledge gained from the pilot study to enhance the software architecture as well as the study design, both of which are described in the relevant sections below.

## 4.2 Participants

100 individuals were recruited for this study via YouGov's online panel. The panel consists of 17 million registered members across 50 markets covering the UK, Americas, Europe, the Middle East and Asia Pacific. Participants for the study were recruited from the Portland, Oregon area since individuals were required to come into a regional center and experience the interaction on standardized hardware. This ensured standardization and integrity of the collected data streams. Study inclusion criteria included: (i) between 24-65 years of age, (ii) native english speaker, (iii) currently employed, and (iv) having at least 3 years of work experience. Exclusion criteria included: (i) having a visual impairment or (ii) having a hearing impairment. Standard procedures were followed for information, informed consent and recording consents for all participants.

*Sample characteristics:* Of the N=51 valid participants (see Section 4.4.1), 59% were female and 41% were male. 92% were employed full time, 8% were self-employed, and the average age was 41.2 years (SD=9.2). The study sample closely mirrored the local population with respect to race and ethnicity based on recent US census data, with participants self-identifying as 84% White, 6% American Indian, 2% Black, 4% Asian, and 4% Latinx. With respect to self-reported total annual income, 14% were between $25,000 to $49,999, 34% were between $50,000 to $99,999, and 52% were $100,000 or more.

## 4.3 VR Software Enhancements

Data from the pilot study revealed a misalignment between the continuous impact that a participant was having on the avatar and the associated body language and facial expressions exhibited by the avatar. The theory of learning to be effective interpersonally requires people to recognize when they are producing a negative effect, so they will know to change to some alternative behavior that may be more useful. The aforementioned misalignment conflated the ability for one to discriminate the impact they were having on the avatar, since human behavior is subconsciously driven by how we perceive social cues (Louwerse, M. M., Graesser, A. C., Lu, S., & Mitchell, H. H., 2005) such as changes in body language and facial expression. As a result, we determined that we did not have the right closed-loop framework to understand human behavior with avatars in a *metaversal learning* context. To alleviate this issue and reduce the cognitive load of the inhabiter from having to map specific Xbox controller-driven avatar behavior to align with the continuous data, we developed a novel audio-based artificial intelligence (AI) algorithm described below and incorporated it into the foundational architecture of the software, described in Section 3.2.

4.3.1 *Audio-based Artificial Intelligence (AI):* To develop the new algorithm, we first recorded and studied a library of videos of individuals interacting with each other in a video-based conferencing system and in recreated natural settings, paying particular attention to the facial expressions, microgestures, body poses and head motion of these individuals. We extracted facial features using the OpenFace library (Baltrusaitis, T., Zadeh, A., Lim, Y. C., & Morency, L. P., 2018) and audio features using the Praat audio library (Boersma, P., 2001) for frontal-facing videos in our library. While the specific details of our technical implementation are considered beyond the scope of this manuscript, we first established correlations between the length and signal characteristics of audio segments and non-verbal behaviors such as facial expressions, body pose, frequency of body pose changes, head motion, and frequency of microgestures among other things during these interactions. We analyzed these correlations when an individual was speaking separately from when an individual was listening during the interaction. This data was used to inform the creation of an animation library using a combination of Optitrack/Motive, Xsens Suit, Manus gloves, and a head mounted GoPro camera system, with nearly 40 minutes of animation. Maya and MotionBuilder were used to create the avatar control rig. Using the previously described correlations between audio segments and nonverbal behaviors, we used a real-time decision tree framework (Brijain, M., Patel, R., Kushik, M. R., & Rana, K., 2014)) to drive the facial expressions, posture changes, gestures, and microgestures of the avatar. A brief video demo of the AI algorithm used to drive an avatar using purely audio signals can be viewed [here](here).

4.3.2 *Closed-loop control and Events of Interest:* In addition to the continuous impact assessments described in Section 3.4.1, the inhabiter was also trained to indicate any points of interest in the conversation with a positive or negative valence. These were separately time stamped and labeled as 4 and 5 respectively using the keyboard interface (page up and page down keys). The inhabiter was allowed to make these valence changes and mark points of interest as often as needed throughout the interaction, based on the performance of the participant. For the reasons described at the start of Section 4.2, we wanted to close the loop between the continuous impact data and events of interest indicated by the inhabiter and the behavioral manifestation of the avatars. This data was therefore used as an input to the decision tree algorithm, providing us the ability to alter decision nodes during traversal. This ensured that the body language and facial expressions of the avatar reflected the valence (positive, negative, or neutral) that the participant was having in the moment. Additionally, the events of interest triggered subtle changes in facial expressions or microgestures that were aligned with the valence of the event, providing immediate contingent feedback to the participant about the interaction.

4.3.3 *Enhanced Lip Synchronization:* The pilot study also revealed that the phoneme-based algorithm we used for lip synchronization (see Section 3.2.1) resulted in rendering artifacts for certain words that were noticeable by a few participants. We chose to enhance this algorithm by utilizing viseme prediction instead of the phonemes. The real time audio input from an inhabiter was represented as mel-spectrograms at 24kHz sampling rate featuring frequencies from 10Hz to 11.66kHz distributed in 93 mel bands. The baseline model was a speaker-independent autoencoder trained in an unsupervised manner to predict phonetic content from the audio data. This baseline autoencoder was phonetically fine-tuned and trained to optimize categorical cross-entropy on per-frame predictions of both phonemes and visemes. Two post-processing heuristics were employed to improve the stability of the results. The first was a moving average over the predicted probabilities and the second was an uncertainty threshold for transitions (set to 0.5). For target predictions that were below an uncertainty threshold on the first 20ms of the transition, the previous strong target was preserved. To evaluate the performance of the proposed approach we employed two well-known metrics (i) multi-class accuracy and (ii) multi-class ROC AUC with one vs. one configuration. The long scope autoencoder model, while more robust, limited the latency of the model to 130ms while a shorter scope, while performing slightly worse, provided an opportunity to lower the latency to about 90ms which was preferable for our real-time application.

4.3.4 *Video playback and review:* The non-authoritative instance was modified to include the ability for individuals to watch their own video (see Section 3.5.2) after the interaction. To do this, the incoming audio stream of the inhabiter was stitched with the render texture from the simulation along with a video feed of the participant on the local machine. The data was stored on separate channels to facilitate analysis. This video was played back through a drop-down screen in the virtual environment. Participants used the up and down arrow keys to indicate their own impact on the avatar while they were watching the recorded video, and this data was recorded with the same values described in Section 3.4.1

All the above enhancements were tested using a rigorous quality assurance process and used to create the final version of the software used by the participants in the research study.

**4.4 Experimental Design & Protocol**

To evaluate our hypotheses, participants completed 4 different conversations with the same avatar in a single session: a screening conversation (Scenario 1), 2 training conversations (Scenarios 2 and 3), and one assessment conversation (Scenario 4). The order of the training conversations (Scenarios 2 and 3) were randomly assigned to minimize the possibility of order effects. Participant improvement in interpersonal effectiveness was assessed by evaluating change in IMPACT and SURVEY using a pretest-posttest design (between Scenarios 1 and 4). The four *Scenarios* used for the study can be found here. Following the screening conversation (Scenario 1), participants were classified as "successful" or "unsuccessful" with respect to achieving the pre-specified outcome. Participants whose Scenario 1 SURVEY scores were $\geq 7$ were defined as "Successful" and those with scores $< 7$ were defined as "Unsuccessful."

The study was conducted at a physical location in Portland where the standardized hardware to experience the interaction was set up. A moderator facilitated the study and was available to answer any questions that participants had. The inhabiter was based in Durham, North Carolina and a peer-peer connection was made between the authoritative and non authoritative instances to initiate the conversation. Following each conversation, participants filled out the SURVEY questionnaire. The inhabiter completed the SURVEY-I and the participant completed the SURVEY-P. At the end of the final assessment conversation (Scenario 4), participants were also asked to fill out a post-participation questionnaire. A link to all the questionnaires used in the study can be found here.

Based on data from the pilot study, we found that the amount of time spent interacting with an avatar influenced the outcome of the conversation. Rather than controlling for this potential confound when evaluating the performance of participants in this study, the duration of each conversation was held constant. Each interaction therefore had to be completed within 6 minutes, which was approximately the average time taken by participants during their interaction in the pilot study. All individuals were then required to review a video of their own conversation and continuously rate their performance using the IMPACT assessment tool. This process resulted in a continuous stream of data reflecting the participant's estimate of their performance during the conversation.

All four conversations were conducted in a single sitting. The total amount of time to complete all 4 conversations, including orientation, training, and time for the participant to review their own conversations, took approximately 75 minutes. Participants were compensated $XX for their participation.

4.4.1 *Usable / Valid Data:*

Out of the 100 recruited participants, 75 completed the baseline survey. Due to the peer-peer nature of the connection, some participants (~6%) experienced networking or technical issues during the interactions. Additionally, there were instances where the participants did not complete the interaction-specific questionnaires (~26%). For the purposes of maintaining the validity of our analyses on the data, we used a stringent rule of excluding any participants where any such issues occurred. Our final data set used in the analysis therefore contains 204 interactions from 51 participants (4 complete uninterrupted interactions and associated survey data for each participant).

**4.5  Data Analysis and Measurement Framework**

4.5.1  Statistical Analysis

Participants with complete data at each conversation were included in the analyses (N=51). All statistical tests were evaluated using SAS Statistical Software, version 9.4 TS Level 1M7 (SAS Institute, Cary, NC). A series of simple Pearson's correlations were conducted via PROC CORR to examine the relationship between IMPACT and SURVEY scores within each scenario. A repeated measures ANOVA was then conducted to examine change in IMPACT scores across between baseline screening (Scenario 1) and post-assessment (Scenario 4). The main independent variables of interest were group ("successful" versus "unsuccessful") and measurement time (4 scenarios) and the group by time interaction. An unstructured correlation structure was used to capture the within-person correlation over time. The regression model was implemented using PROC MIXED and the Ken-Warl-Roger option was used to obtain the correct denominator degrees of freedom for the F-tests. Residual error terms were assumed to follow a mean-0, normal distribution. The fitted model was used to report average IMPACT scores within each level of the independent variables and to make inferences about within-in and between-group differences across the scenarios. All tests were 2-sided and $p < .05$ was considered statistically significant. The adaptive step-down Bonferroni adjustment (as implemented in PROC MULTTEST) was used to control the overall (family-wise) error rate of all unplanned comparisons. Finally, PROC FREQ was used to conduct a 2 x 2 Chi-Square Goodness of Fit Test to determine whether the proportion of participants rated as "successful" versus "unsuccessful" changed significantly between Scenario 1 and 4.

4.5.2  Overall architecture and collected data streams

Figure 2 is a schematic illustration of the architecture between the software instances used to instantiate the MLE. All participants (non-authoritative instances) were located in Portland, Oregon while the inhabiter (authoritative instance) was located in Raleigh, North Carolina.

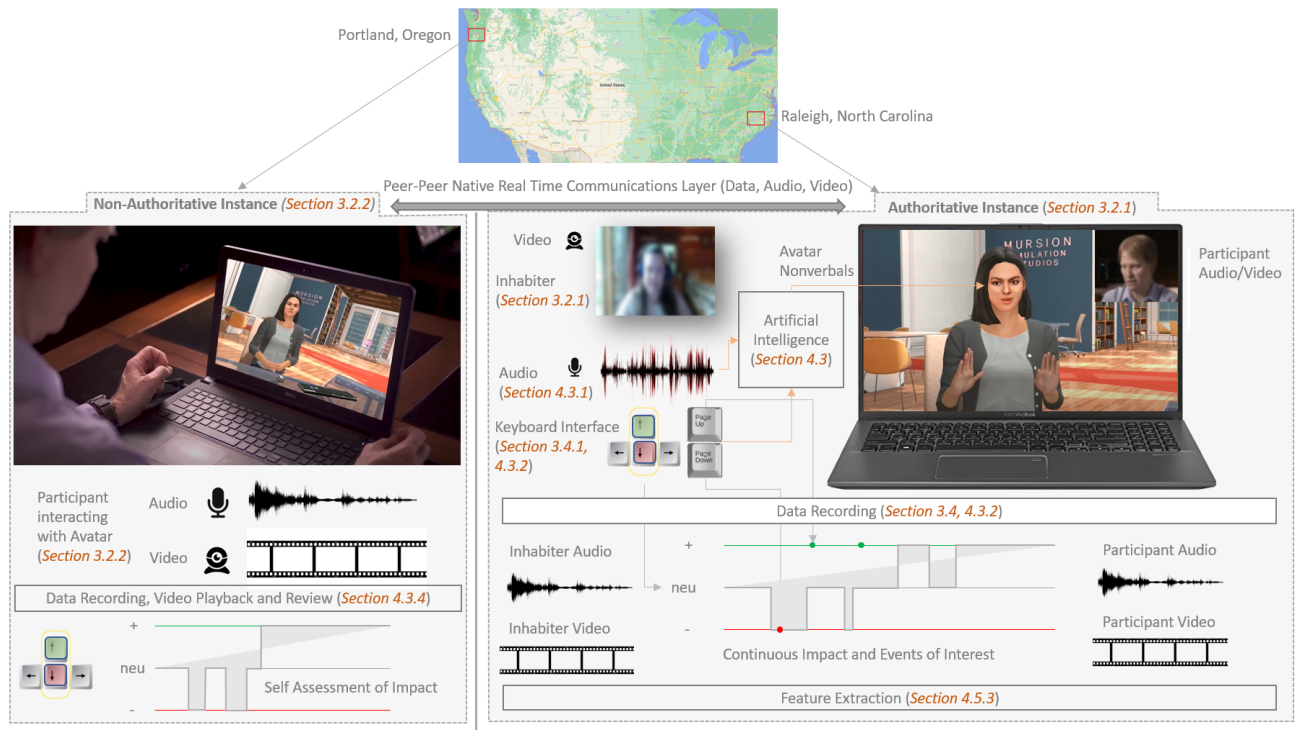

*Figure 2: Schematic showing the overall architecture of the MLE and the data streams produced and collected by each software instance. Shown in brackets (Section) are the relevant sections of the manuscript that contain more details about the relevant architectural components.*

| Data Stream | Description / Source | Sampling / [Resampling] |
|---|---|---|
| Participant Audio (Section 3.4.2) | Spoken dialogue from the participant using a Jabra 2400 Biz Duo II noise-canceling mic and headset | 48 KHz |
| Participant Video (Section 3.4.3) | Video feed of the participant from a 1080p NexiGo N60 USB Camera | 29.79 Hz |
| Inhabiter Audio (Section 3.4.2) | Spoken dialogue from the inhabiter using a Jabra 2400 Biz Duo II noise-canceling mic and headset | 48 KHz |
| Inhabiter Video (Section 3.4.3) | Video feed of the inhabiter from a Logitech 720p USB webcam | 15 Hz |
| *The following data streams were originally collected via an event-based system but resampled to match the render frame rate of 30 frames per second for analysis:* | | |
| IMPACT (Section 3.2.1) | Inhabiter's assessment of the impact that the participant was having on their avatar | [30 Hz] |
| Events of Interest (Section 4.3.2) | Inhabiter's indication of a discrete event of interest that occurred during the interaction, | [30 Hz] |
| Self Assessment (Section 4.3.4) | Participants' self assessment of the impact they were having on their avatar | [30 Hz] |

*Table 1: Details of the various data streams collected during the human-avatar interaction in the metaversal learning environment.*

### 4.5.3 Feature Extraction in Video and Audio Streams

Since the MLE framework presented here provides us with a rich stream of continuous data (Section 4.5.2), we analyzed these streams to discover any correlations between the underlying data and the final outcome of the interaction between the participants and the avatar. We developed a web-based interface (Figure 3) to verify the validity and integrity of all the collected data streams, allowing us to alter the thresholds used in the algorithms for voice-activity-detection, identify synchronization issues, and remove interactions that contained missing data from the analysis. Our goal was to build a machine learning classifier capable of predicting whether participants were successful or unsuccessful in their interaction with the avatar based on the data gathered during these conversational scenarios. Two primary sets of features were extracted from the collected data.

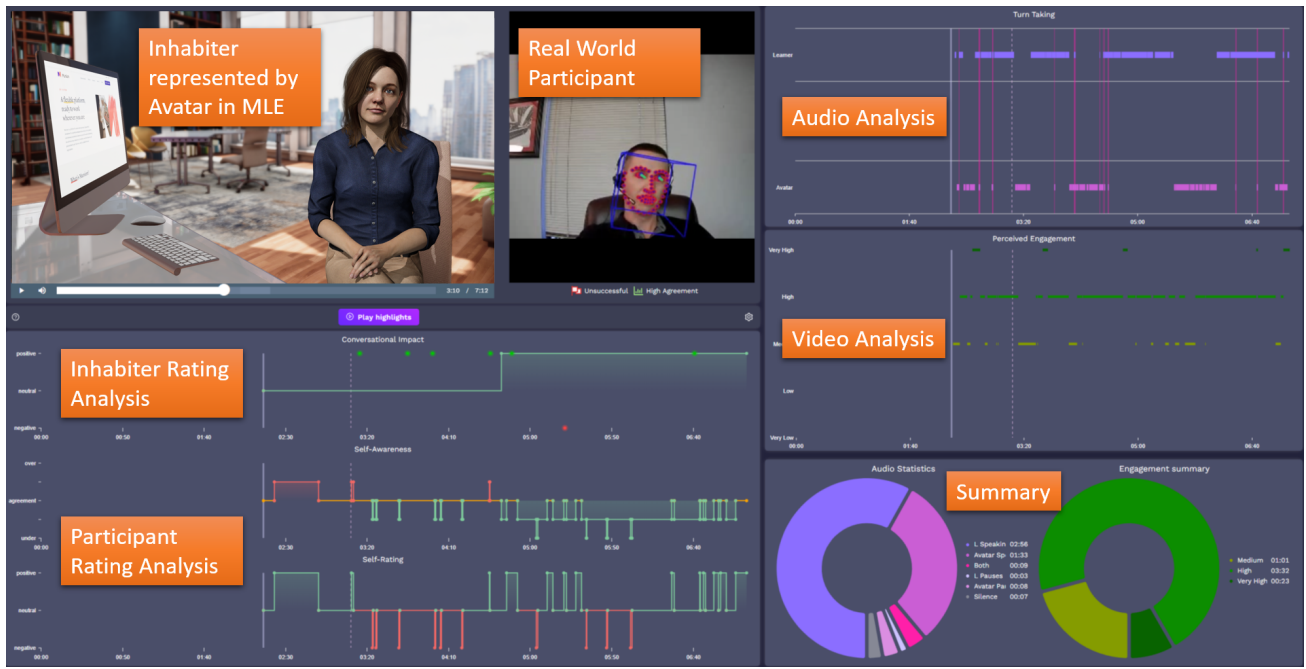

*Figure 3: A web-based interface was developed to perform synchronized playback of the recorded data and verify the integrity and validity of all the collected data streams from the 204 simulations, prior to training the machine learning models.*

4.5.3.1 Video features: For the video streams recorded, a 'video feature set' was created using pre-trained machine learning models, specifically, OpenFace (OF) (Baltrusaitis, T., Zadeh, A., Lim, Y. C., & Morency, L. P., 2018) and an emotion recognition model (Hume FMM, 2022). First, a participant's video was processed with OF: for each frame of video extracted, the face was detected and cropped; then, an enhanced pretrained model for emotion recognition (Cowen, A. S., & Keltner, D., 2020) detected a set of 48 emotions for each frame. Based on the work by (Ahn, Junghyun & Gobron, Stéphane & Silvestre, Quentin & Thalmann, Daniel. 2010 (see Fig 2. *Two dimensional circumplex space model and its emotional sample*)), these 48 emotions were mapped onto a 2D space (or 2D map), with each emotion characterized as either active or passive, and either positive or negative. Coordinates for each emotion were interpreted from the 2D space, and emotions were grouped into 8 clusters: active-positive, active-negative, passive-positive, strongest-passive-positive (background emotions like 'calmness' and 'concentration'), passive-negative, and three other clusters for the perceived engagement of a participant (see Appendix A for details). For each frame of video an emotion vector was calculated as

$$v = \sum_i probability_i \cdot coordinates_i$$

A comparison of the clusters belonging to two representative subjects on either end of the successful and unsuccessful spectrums of performance revealed a significant disparity in the footprint of the emotion vectors (see *Figure 4*). This provided the basis for us to train a classifier based on the features obtained by the emotion recognizer. Principal Component Analysis (PCA) was done based on the emotion vectors, and the center-of-mass was calculated for each session.

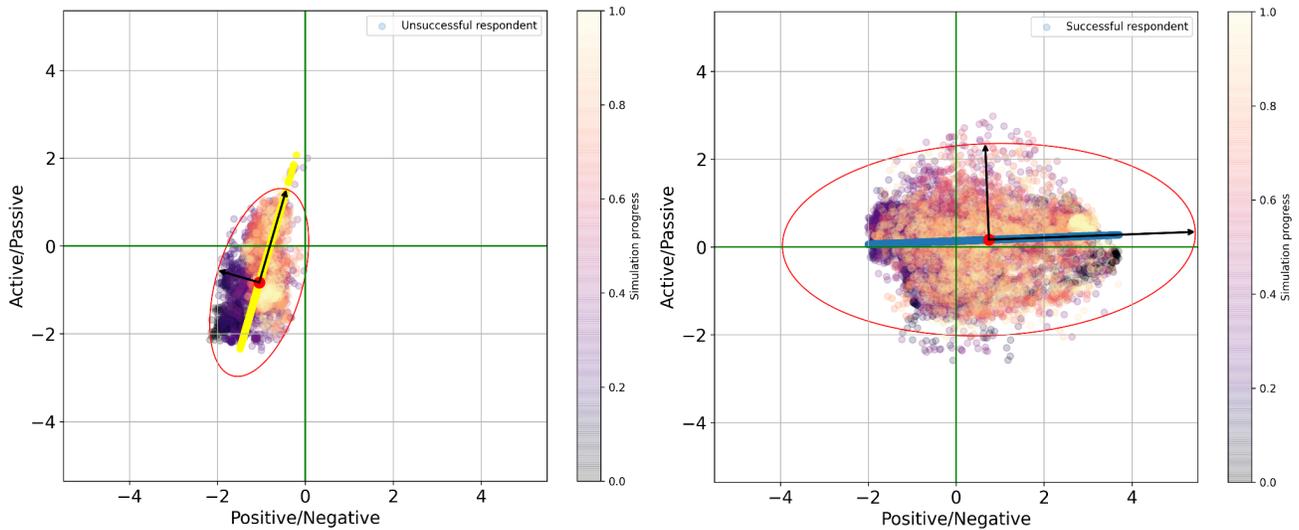

*Figure 4: <u>(Left) Low Interpersonal Effectiveness:</u> Probabilities of extracted facial features perceived by an emotion detection algorithm are concentrated in the negative region, and mostly in the passive quadrant. The center of mass is located in the passive-negative region. <u>(Right) High Interpersonal Effectiveness:</u> The probabilities of extracted facial features perceived by an emotion detection algorithm for the entire interaction are well-distributed over the quadrants, with their center of mass located in the active-positive region. Also shown in both images are the confidence ellipsis, centers-of-mass, main components of Principal Component Analysis (PCA) decomposition, and PCA-explained-variance-vectors. Since these features were extracted for a video sequence, time (normalized) is indicated as color and accounts for each frame of processed video. The color therefore indicates the change in detected emotional expression for a participant during a simulation. For example, for the individual with a low interpersonal effectiveness score (Left), we can see that at the start of the interaction, perceived emotions were mostly negative (dark blue cluster centered around [-1.7, -1.7], but by the end of the simulation, their perceived emotions became slightly **more** positive (yellow cluster centered at [0.8; -1.2]).*

4.5.3.2 <u>Audio features:</u> The second set of features, an 'audio feature set', was taken from the participants' audio with a version of the Praat software (Jadoul, Y., Thompson, B. and De Boer B., 2018). Each audio file was processed to reduce noise, and silences in the audio segments were removed. Three subsets of features were then extracted from each audio segment. The first subset consists of basic audio statistics, like mean, median, minimum, and maximum values of the fundamental frequency *F0* (Boersma, Paul., 2001), jitter (variation in periods), shimmer (cycle-to-cycle variation in intensity), harmonicity (degree of acoustic periodicity, also called Harmonics-to-Noise Ratio (HNR)), mean and median values of the first four formants (frequency peaks in the spectrum which have a high degree of energy, especially prominent in vowels; each formant corresponds to a resonance in the vocal tract). All these features were calculated using a moving window of 0.5 seconds and no overlap, with the final feature set having a mean value calculated across all these windowed segments. The second subset included the average formant, formant dispersion, interaction of glottal-pulse rate and vocal-tract length in judgements of speaker size, sex, and age, vocal tract length, and formant spacing. The third subset consisted of an overall snapshot of the interaction including duration of speech, number of syllables, phonation time, number of pauses, speech rate (number of syllables divided by speech duration), and articulation rate (number of syllables divided by phonation time).

The extracted audio and video features were analyzed holistically, as well as individually to determine if they had any predictive power in line with our hypothesis, while also allowing us to determine the computational complexity and burden of utilizing those features in an informative manner.

### 4.6 Results

4.6.1 *Is there a relationship between Continuous Ratings of Interpersonal Effectiveness (IMPACT) and Post-Conversation Ratings of Success (SURVEY)?* To evaluate the relationship between IMPACT and SURVEY ratings, separate Pearson's correlations were conducted between these two measures at each scenario. The correlation was 0.69 ($p < .0001$) for Scenario 1, 0.13 ($p = .3446$), 0.69 ($p < .0001$) for Scenario 3, and 0.49 ($p < .0002$) for Scenario 4. With the exception of Scenario 2, these findings suggest that those participants who were rated as more likely to achieve the Scenario goal (SURVEY) were also more likely to be interpersonally effective (IMPACT) during the conversation.

4.6.2 *The relationship between Continuous Ratings of Interpersonal Performance (IMPACT), practice, and Post-Conversation Rating of Success (SURVEY):* Results from the repeated measures ANOVA reveal a significant main effect for group (SUCCESS), $F(1, 49) = 13.49$, $p < .0006$ and a significant time X group interaction (SUCCESS X SCENARIO), $F(1, 49) = 6.72$, p < .0125. The main effect for time (SCENARIO) was not statistically significant ($p = .5597$. The significant main effect for the group (SUCCESS) suggests that the average IMPACT scores across conversations was significantly different between "successful" and "unsuccessful" participants. The predicted average mean score for "successful" participants was 117.28 (SE = 11.00) and 62.77 (SE = 9.96) for "unsuccessful" participants: a difference of 54.51 (SE = 14.84) points. However, the significant interaction between group and time (SUCCESS X SCENARIO) suggests that the average within-in group change between conversation 1 and 4 varied by success status. Planned pairwise between Scenario 1 and 4 within each level of success status revealed that "unsuccessful" participants had significantly higher average IMPACT scores at Scenario 4 compared to Scenario 1 (Scenario 1 M = 46.15 (SE = 14.26) and Scenario 4 M = 79.39 (SE = 9.68), a difference of 33.24 (SE = 14.04) points, $p < .0219$. Although participants rated as "successful" at Scenaior 1 had lower average IMPACT scores at Scenario 4 (Scenario 1 M = 127.76 (SE = 15.74) and Scenario 4 M = 79.39 (SE = 9.68), this difference was not statistically significant (p = .1822). Figure X presents the average IMPACT scores along with their standard errors at each time point by success status and relevant p-values for each pairwise comparison between the screening conversation (Scenario 1) and the assessment conversation (Scenario 4). Between-group comparisons at each scenario found that at Scenario 1 "unsuccessful" participants had significantly lower IMPACT scores on average than "successful" participants, by Scenario 4 this difference was not statistically significant (p = .0633).

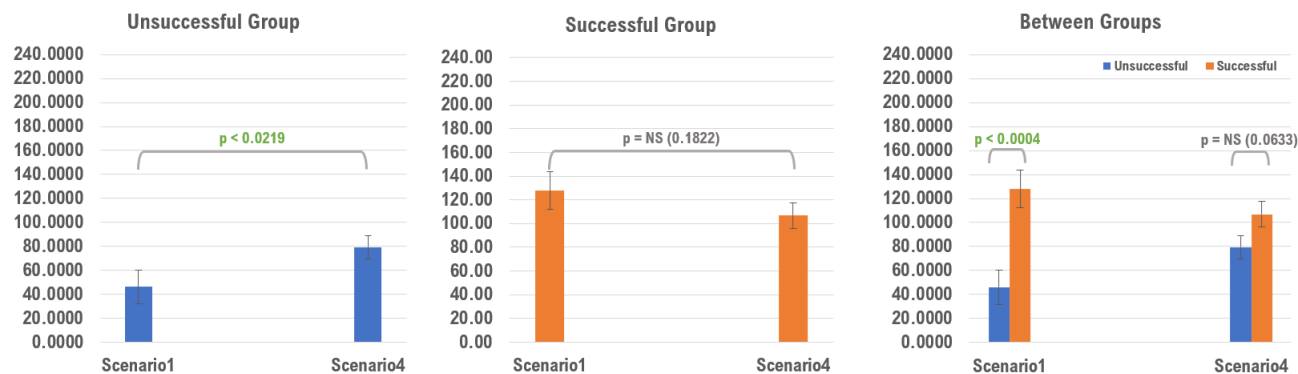

*Figure 5: Individuals who were identified as having deficits after the baseline Scenario (Unsuccessful Group) showed a significant improvement by Scenario 4. The difference in interpersonal effectiveness for the Successful group between Scenario 1 and Scenario 4 was not significant. A between group comparison reveals that the unsuccessful group after Scenario 1 showed no significant differences from the successful group after Scenario 4.*

4.6.3 *Does the proportion of those rated as successful change from Scenario 1 to Scenario 4?*

Following the baseline screening, 23 (45.1%) of participants were rated as "successful" and 28 (54.9%) were rated as "unsuccessful." By the end of Scenario 4, 32 (62.75%) participants were rated as "successful" and 19 (37.25%) were rated as "unsuccessful". More importantly, 12 (42.86%) of the baseline "unsuccessful" participants successfully achieved the Scenario 4 goal, while only 3 (5.88%) of those rated as "successful" at baseline failed to achieve the Scenario 4 goal. The proportion of those improving by *Scenario* 4 was significant, $\chi^2$ (1, N= 51) = 10.51, p < .0012. Similar to the findings of change across time on average IMPACT scores, those rated as "unsuccessful" at *Scenario* 1 are more likely to change success status categories by *Scenario* 4.

4.6.4 *Can verbal and nonverbal data be used to predict Conversational Success?*

Video and audio feature sets were computed for 204 simulations (4 different scenarios per respondent, 51 respondents in total) as described in the section *4.5.3*. Out of 204 simulations, 130 had a positive outcome ('pass', ~64%). Support vector machine with a linear kernel was chosen as a classifier and all results are computed using a leave-one-out cross-validation methodology. We built several classifiers using: (1) the full audio and video feature set, (2) using only audio data, (3) using only video data, and (4) using only a subset of the features (top features) that yielded results that did not compromise the overall performance.

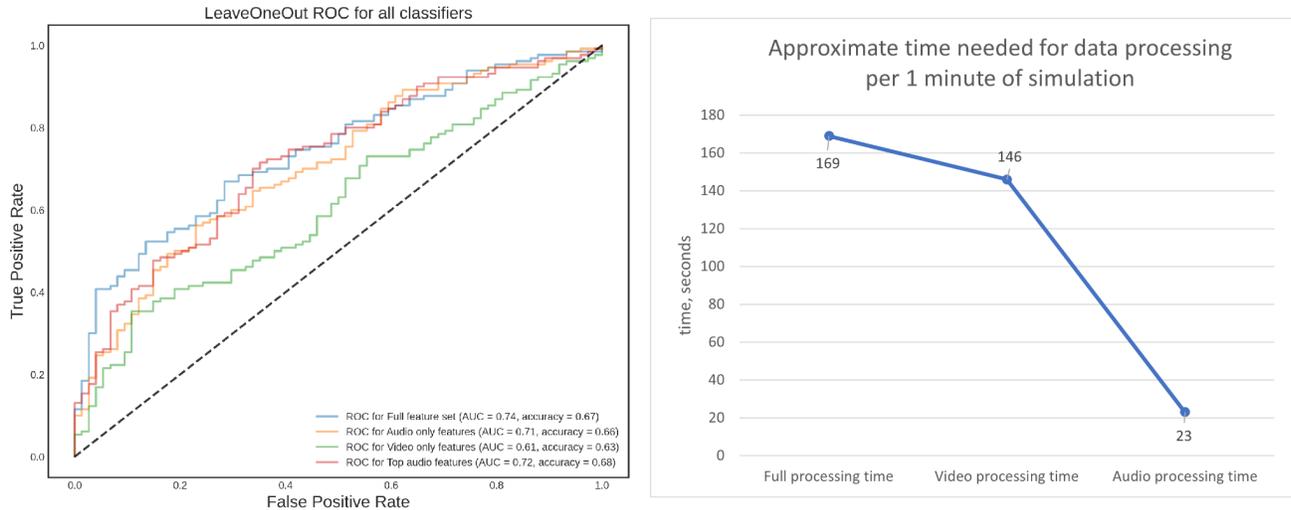

*Figure 6: Left: ROC curves for the SVM trained on full feature set, audio only, video only, and top selected features. Right: Approximate time needed for data processing per one minute of simulation, in seconds. By removing video features, computing time was reduced by 85%.*

As seen from the ROC curves (see *Figure 6*), the audio-only classifier gives very similar accuracy to that built on the full feature set. In order to decrease computational complexity, make the classifier results easier to interpret, and remove any possible security and privacy concerns associated with video streams (Kagan, D., Alpert, G. F., & Fire, M., 2020), we tried to find a minimum set of audio features so that the classifier's accuracy would be very close to that trained on a full feature set. It was done in two steps: first, we took the top-20 features by feature importances (coefficients (weights) assigned to each feature by the SVM). Then we compute a correlation matrix of these top-20 features, and remove several with a correlation higher than 0.9. By doing so, we've reduced the number of features by three times, from 53 to 17. To check that classifiers don't overfit, we've performed the Leave-One-Out cross-validation, where each sample is used once as a test set, and all the remaining samples are used as a training set. As seen in *Fig.* <>, area under the curve (AUC) for the classifier trained on top selected audio features is ~0.72 and an accuracy of 68%, which is very similar to both 'full' and 'audio only' classifiers. This is a promising result given that we only consider a few audio features averaged over the entire duration of the interaction and without any information on the dynamics of the conversation. To put this in perspective, an accuracy of 77% percent was achieved using video-audio non-linguistic features that included turn taking information (Pentland, A., & Heibeck, T. 2008, Byun, B., et.al. 2011). Besides, this feature selection reduced the computing time needed for data preparation by ~85%, while the accuracy and AUC of such classifier have a difference of only 1-2%. *Figure 6* above also shows approximate time needed for data preparation in all three cases: full feature set, video only, and audio only.

## 5 General Discussion

The results described in the previous section highlight some important aspects that can be useful in creating highly effective *MLEs* to measure interpersonal effectiveness.

### 5.1 Self Awareness

After participants watched a video of their interaction and rated their own continuous impact (see Section 4.3.4), they were administered a post simulation survey that asked them to assess whether or not they achieved the intended conversational outcome on a scale of 1 to 10. This was compared to the inhabiter's assessment of the participant's performance using the same survey scale. We considered participants whose assessment of their own performance was within a single point of the inhabiter's assessment of their performance to be accurate-estimators, or being "self-aware" of their performance. Similarly, over-estimators, and under-estimators, were identified depending on the direction in which their assessment differed from that of the inhabiter by 2 or more points on the scale. Figure 7 shows the results by accounting for floor/ceiling effects during this analysis.

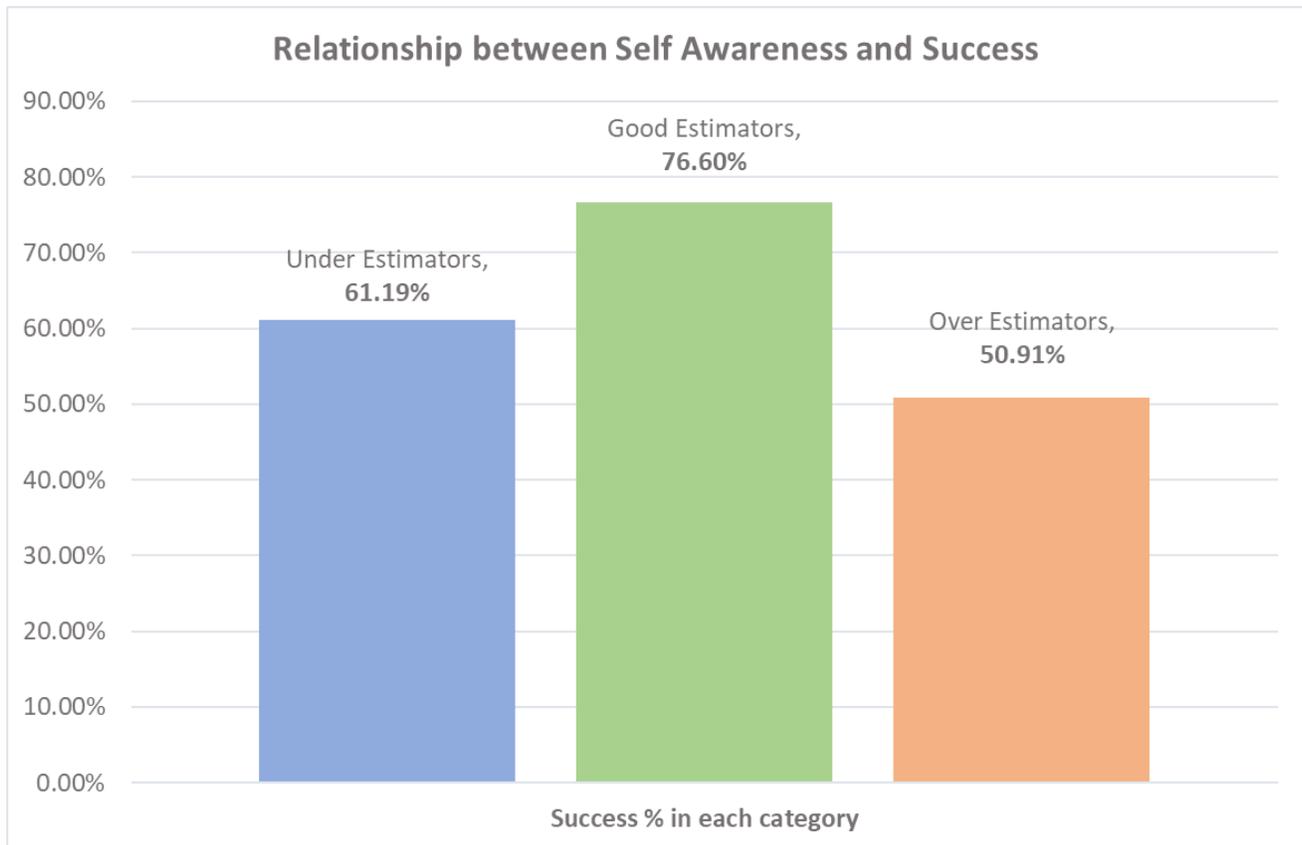

*Figure 7: Data where the inhabiter's assessment of the participant's performance was < 2.5 or > 8.5 was eliminated from the analysis to account for floor/ceiling effects of the 1-point differential. Analysis of the percentage of participants that were successful in each of the three categories revealed that over-estimators or under-estimators were less likely to be successful than good-estimators.*

In total, 66% of participants were successful in their interaction with the avatar across all the simulations. Of these, ~78.5% were accurate-estimators or under-estimators. Of the total number of over-estimators, only ~50% were successful in their interaction with the avatar. In other words, participants who are self-aware of their own interpersonal effectiveness were much more likely to be successful during the interpersonal interaction than their less self-aware counterparts.

## 5.2 Stimulus properties of Avatar and Scenarios

Results revealed that the interaction between the participants and the avatar in the MLE produced results that we would have expected when individuals interact with each other in the real world. We attribute this similarity in results to the stimulus properties of the avatar and the scenarios. In order to allow individuals to interact with an avatar in a manner that they would have done with another person in real life, the appearance, verbal and nonverbal responses of the avatars needed to have believability and not detract from the experience. We believe that the advances to the VR software and AI algorithms (see Section 4.3) allowed an inhabiter to effectively facilitate a natural human-avatar interaction. The results also indicate that the *Scenarios* which set the context of the conversation had properties that allowed the participants to interact with the avatar as they would with an individual in the real-world. Both of these lead us to believe that the presented framework reinforces the fundamental concepts of Situational Plausibility and Place Illusion (Slater, M., 2009) required to create realistic behavior in MLEs.

### 5.3 Generalization and Repeatability

Our analysis has shown a correlation between interpersonal effectiveness and success in achieving a conversational outcome, while also highlighting that interpersonal effectiveness can be learned and improved with repeated interactions with an avatar. The improvement, however, was seen only among those who rated as "unsuccessful" at the screening conversation. This suggests that MLEs can be used to screen and target training to those who will more likely benefit from repeated interactions. While these results may include practice effects, they suggest that interpersonal effectiveness can be learned and improved across various conversational contexts. To maximize generalization, MLEs provide the affordance to vary the characteristics of both the avatar and the scenario given the stimulus properties described in Section 5.2. We plan to investigate this in our future work. In summary, results indicate that the effectiveness and outcome of interpersonal interactions between avatars and humans in MLEs is measurable through multi-exemplar training using the framework described in this manuscript.

### 5.4 Predicting Interpersonal Effectiveness

Our results confirm that MLEs offer the ability to collect high fidelity datastreams which can augment our understanding of interpersonal interactions. Preliminary results of processing the audio and video from the interaction between participants and the avatar show promise in being able to predict conversational success. At this stage, we trained our predictive algorithms on features that were extracted holistically for participants that were "successful" vs. those that were "unsuccessful". These algorithms could be further refined by extracting features in specific segments of a conversation independent of whether or not that individual was "successful" or "unsuccessful" using the IMPACT score framework described in Section 3.4.1.2. Such a self-supervised learning approach will become increasingly powerful as new data collection pipelines allow thousands of conversations to be analyzed over time.

We believe that such a framework can be used to provide real-time cues to individuals as a conversation in the real-world is progressing, providing them with an opportunity to alter their behavior to ensure conversational success. The applications for such frameworks are many including sales, customer service, clinical interventions, and other professional environments.

We also foresee using these data to investigate fundamental parameters of human perception. For example, information about when users notice positive and negative events during a conversation can be used along with the survey ratings to infer how impactful and sticky negative events are compared to positive ones. Our preliminary investigations in this area are promising and a complete exploration is reserved for future work.

## 5.5 Self-Reported Measures

In addition to the data collected in the MLE, all participants were asked to complete a pre-experience (ISES) and post experience questionnaire that would help us understand how they perceived the overall interaction with the avatars. Prior to the interaction, factor analysis (Cronbach's Alpha = 0.85) revealed that less than 10% of the participants felt that they had the skills to be interpersonally effective. Following the series of interactions, the following questions were administered. The results are presented below.

*Q1. If you had the opportunity to repeatedly practice your communication skills in this type of environment, how much skill improvement do you think you would see over time ?*

34.48% of the participants thought they would see an extremely significant improvement while 58.62% thought they would see a significant improvement. In summary, 93.1% of the participants self-reported that they are likely to see an improvement in their communication skills with repeated practice with an avatar in an MLE.

*Q2. I feel confident about engaging in one-on-one conversations around topics that include providing difficult or awkward feedback to a coworker.*

28.81% of the participants strongly agreed with this statement and 55.93% somewhat agreed with this statement. When compared to the results from the ISES survey (<10% indicating they considered themselves to be interpersonally effective), this result where 84.74% of the participants felt somewhat confident in their one-on-one conversation abilities highlights the possibilities for learning in MLEs.

## 5.6 Limitations and Planned Future Work

Our study has several limitations. First, the relatively small homogenous sample for the current study did not allow for an investigation of potential moderators of outcome, such as gender, age, race, years of work experience or professional role. Future studies with larger sample sizes and a more heterogeneous sample, including employees with documented work interpersonal difficulties will be key to determining the usefulness of this intervention. Second, it is difficult at this point in our programmatic line of research to separate practice effects (defined as improvement across time due to increased familiarity with apparatus) from learning effects. However, the finding that improvement was not universal but limited to only those rated as "unsuccessful" at screening provides initial support that improvement was the result of learning rather than familiarity. Third, the surprising finding that average IMPACT scores was lower for Scenario 2 in both groups suggests that some Scenarios may present different learning challenges for participants or may be more difficult for inhabiters to evaluate. In future work, we plan to use Scenarios with normative data to evaluate this theory.

# 6 Conclusion

The primary finding of this research is that avatars can be used to help individuals improve their interpersonal effectiveness. Specifically, individuals with deficits show significant improvements in their interpersonal effectiveness skills by repeatedly interacting with avatars in MLEs. These interactions can be designed to simulate situations that occur in professional or personal settings. Specifically, the stimulus properties of the avatar and scenario can be easily altered to vary the learning challenge or tailor these to be more appropriate to the learning context. As a part of our research, we developed a quantitative framework to consistently measure interpersonal effectiveness using avatars. A single *Scenario* (Section 3.3) within this framework can be used to help classify individuals into groups and help identify individuals with deficits. Additional *Scenarios* can then be personalized for those showing deficits and used to help improve their interpersonal effectiveness. Our research demonstrates that individuals who were categorized as having a deficit initially, showed an improvement in interpersonal effectiveness after only 3 additional *Scenarios*, each lasting approximately 15 minutes, including the time taken to review a video of their own conversation with the avatar. This has implications for professional learning and development, since MLEs can be used to improve interpersonal effectiveness and performance of the workforce in almost any vertical including customer service, sales, leadership, education, and healthcare, among others. Further research is needed to establish the dose, timing, and frequency of such interactions in an MLE that will help individuals acquire and retain skills in their repertoire to remain interpersonally effective in their personal and professional careers.

In addition to the main finding of our research, preliminary data from the main study suggests that self-awareness, or recognizing one's impact accurately, may be a component of effective interpersonal behavior. By utilizing avatars to practice complex interpersonal conversations and training individuals to be better at estimating their impact during the conversation, we believe that individuals can learn to be interpersonally effective. We plan to investigate this in our future work. Finally, we analyzed the rich underlying data streams that were collected during each human-avatar interaction within the MLE. While it is still early to draw conclusions, results suggest that the perceived verbal and nonverbal behaviors of individuals during interpersonal conversations may play a role in determining their interpersonal effectiveness. This theory was directionally supported by the results obtained when we trained a machine learning algorithm to predict interpersonal effectiveness. Our future work will involve continuing to refine these models to improve predictive accuracy. To conclude, our research highlights that interacting with avatars in MLEs offers an effective way to practice challenging conversations in a safe and scalable way. We conclude that avatars can help measure, predict, and improve interpersonal effectiveness.

# Appendix A. Mapping Emotions on to a 2D space

All 48 emotions that were detected by a pretrained machine learning model (Hume FMM, 2022) were mapped onto a 2D map, based on the work by (Ahn, Junghyun & Gobron, Stéphane & Silvestre, Quentin & Thalmann, Daniel. 2010 (see Fig 2. *Two dimensional circumplex space model and its emotional sample*)). Each emotion was characterized as either active or passive, and either positive or negative. Coordinates for each emotion were interpreted from the 2D space.

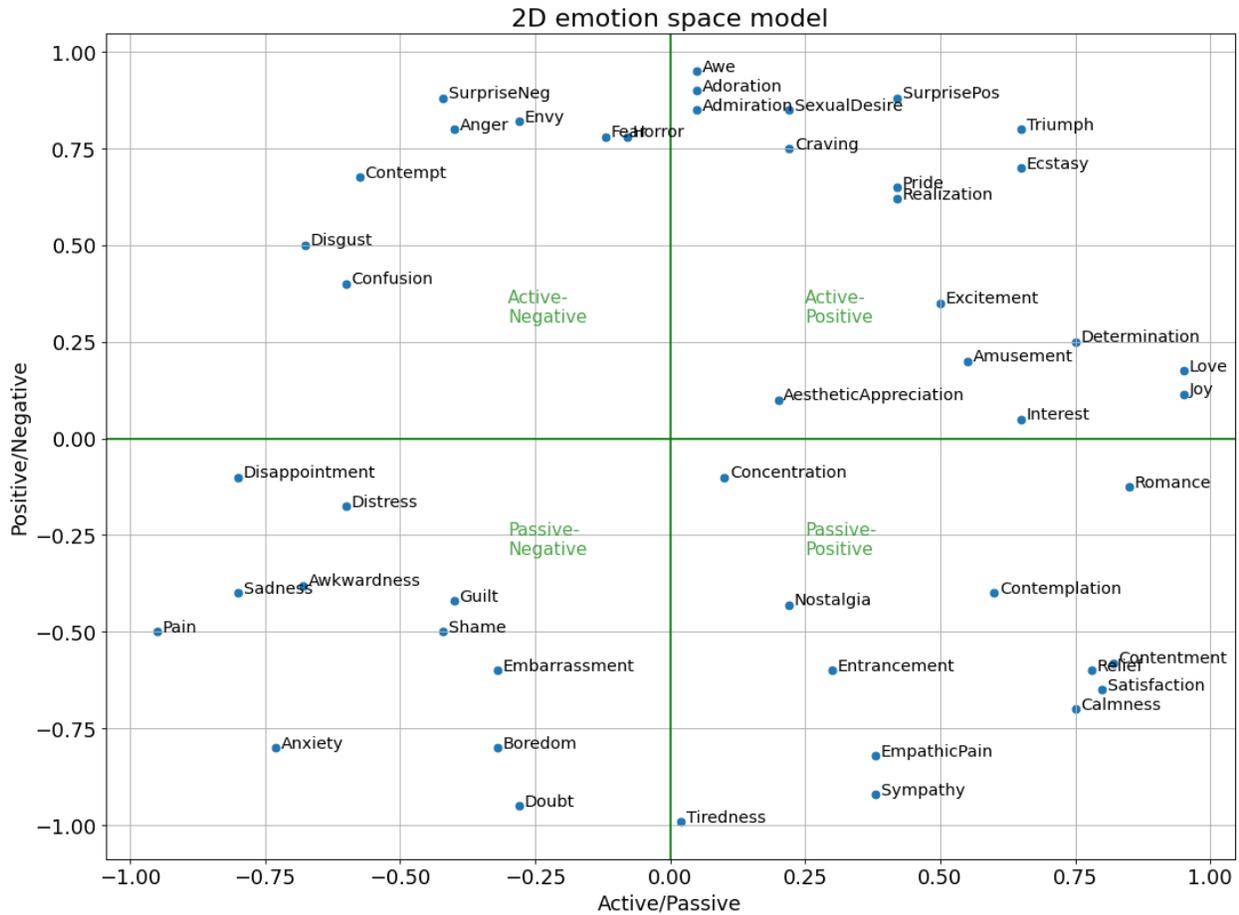

*Figure 8. 2D emotion space map (all 48 detected emotions shown).*

| Emotion | *x* | *y* | Emotion | *x* | *y* |
|---|---|---|---|---|---|
| Admiration | 0.05 | 0.85 | Entrancement | 0.3 | -0.6 |
| Adoration | 0.05 | 0.9 | Envy | -0.28 | 0.82 |
| Aesthetic Appreciation | 0.2 | 0.1 | Excitement | 0.5 | 0.35 |
| Amusement | 0.55 | 0.2 | Fear | -0.12 | 0.78 |

| Emotion | X | Y | Emotion | X | Y |
|---|---|---|---|---|---|
| Anger | -0.4 | 0.8 | Guilt | -0.4 | -0.42 |
| Anxiety | -0.73 | -0.8 | Horror | -0.08 | 0.78 |
| Awe | 0.05 | 0.95 | Interest | 0.65 | 0.05 |
| Awkwardness | -0.68 | -0.38 | Joy | 0.95 | 0.115 |
| Boredom | -0.32 | -0.8 | Love | 0.95 | 0.175 |
| Calmness | 0.75 | -0.7 | Nostalgia | 0.22 | -0.43 |
| Concentration | 0.1 | -0.1 | Pain | -0.95 | -0.5 |
| Contemplation | 0.6 | -0.4 | Pride | 0.42 | 0.65 |
| Confusion | -0.6 | 0.4 | Realization | 0.42 | 0.62 |
| Contempt | -0.575 | 0.675 | Relief | 0.78 | -0.6 |
| Contentment | 0.82 | -0.58 | Romance | 0.85 | -0.125 |
| Craving | 0.22 | 0.75 | Sadness | -0.8 | -0.4 |
| Determination | 0.75 | 0.25 | Satisfaction | 0.8 | -0.65 |
| Disappointment | -0.8 | -0.1 | Sexual Desire | 0.22 | 0.85 |
| Disgust | -0.675 | 0.5 | Shame | -0.42 | -0.5 |
| Distress | -0.6 | -0.175 | Surprise (positive) | 0.42 | 0.88 |
| Doubt | -0.28 | -0.95 | Surprise (negative) | -0.42 | 0.88 |
| Ecstasy | 0.65 | 0.7 | Sympathy | 0.38 | -0.92 |
| Embarrassment | -0.32 | -0.6 | Tiredness | 0.02 | -0.99 |
| Empathic Pain | 0.38 | -0.82 | Triumph | 0.65 | 0.8 |

*Table 2. List of all 48 emotions and their coordinates in a 2D space.*

Based on the mapping, all emotions were grouped into 8 clusters: active-positive, active-negative, passive-positive, strongest-passive-positive (background emotions), passive-negative, and three other clusters for the perceived engagement of a participant. Emotion vectors were created based on the same 2D map.

| Cluster | Emotions |
|---|---|
| Active-Positive | Amusement, Craving, Determination, Ecstasy, Excitement, Joy, Love, Pride, Satisfaction, Sexual Desire, Surprise (positive), Triumph, Interest, Realization |
| Active-Negative | Anger, Contempt, Disgust, Distress, Confusion, Embarrassment, Empathic Pain, Fear, Horror, Pain, Envy, Guilt, Surprise (negative) |
| Passive-Positive | Admiration, Adoration, Aesthetic Appreciation, Awe, Contentment, Entrancement, Nostalgia, Relief, Romance, Sympathy |
| Strongest-Passive-Positive (Background emotions) | Calmness, Contemplation, Concentration |
| Passive-Negative | Anxiety, Awkwardness, Boredom, Disappointment, Doubt, Sadness, Shame, Tiredness |
| Engaged-Positive | Determination, Excitement, Joy, Satisfaction, Surprise (positive), Interest, Realization |
| Engaged-Negative | Disappointment, Surprise (negative), Anger, Sadness |
| Engaged | Engaged-Positive + Engaged-Negative |

*Table 3. List of all clusters used for analysis.*